%% file: main.tex
\documentclass[journal]{IEEEtran}

\usepackage{amsmath}
\usepackage{amssymb}
\usepackage{graphicx}
\usepackage{booktabs}
\usepackage{siunitx}
\usepackage{cite}
\usepackage{url}
\usepackage[hidelinks]{hyperref}
\usepackage{algorithm}
\usepackage{algorithmic}
\usepackage{subcaption}

\begin{document}

\title{Fast Surrogate Learning for Multi-Objective UAV Placement in Motorway Intelligent Transportation System}

\author{Weian~Guo, Shixin~Deng, Wuzhao~Li, and Li~Li%
\thanks{Weian Guo and Shixin Deng are with the Sino-German College of Applied Sciences, Tongji University, Shanghai, China (e-mail: guoweian@tongji.edu.cn). Li Li (Corresponding author) is with the School of Electronics and Information Engineering, Tongji University, Shanghai, China (e-mail: lili@tongji.edu.cn). Wuzhao Li (Corresponding author) is with Wenzhou Polytech University, Zhejiang, China. (e-mail:lwzhao055@wzpt.edu.cn)}%
}%

\markboth{IEEE Transactions on Intelligent Transportation Systems}%
{Guo \MakeLowercase{\textit{et al.}}: Fast Surrogate Learning for Multi-Objective UAV Placement in Motorway ITS}

\IEEEtitleabstractindextext{%
\begin{abstract}
We address multi-objective unmanned aerial vehicle (UAV) placement for motorway intelligent transportation systems, where deployments must balance coverage, link quality, and UAV count under geometric constraints. We construct a reproducible benchmark from highD motorway recordings with recording-level splits and generate Pareto-optimal labels via NSGA-II. A preference rule yields deployable targets while preserving multi-objective evaluation. We train fast surrogate models that map unordered vehicle positions to UAV count and continuous placements, using permutation-aware losses and constraint-regularized training across set-based and sequence-based architectures. The evaluation protocol combines Pareto quality metrics, success-rate curves, runtime benchmarks, and robustness studies, with uncertainty quantified by recording-level bootstrap. Results indicate that permutation-invariant set models provide the strongest coverage--SNR--count trade-off among learned predictors and approach NSGA-II quality while enabling real-time inference. Under shared budgets, they offer a more favorable success--latency trade-off than heuristic baselines. The benchmark, splits are released to support reproducible ITS deployment studies and to facilitate comparisons under shared operational budgets.
\end{abstract}

\begin{IEEEkeywords}
unmanned aerial vehicles, intelligent transportation systems, UAV placement, multi-objective optimization, surrogate learning, coverage, signal-to-noise ratio.
\end{IEEEkeywords}}

\maketitle

\IEEEdisplaynontitleabstractindextext
\IEEEpeerreviewmaketitle

\input{sections/introduction}
\input{sections/related_work}
\input{sections/problem_formulation}
\input{sections/method}
\input{sections/experiments_results_discussion}

\input{sections/conclusion}

\bibliographystyle{IEEEtran}
\bibliography{references}

\end{document}

%% file: sections/introduction.tex
\section{Introduction}
\label{sec:introduction}
Motorway ITS increasingly relies on timely situational awareness and communication support for tasks such as incident monitoring, temporary congestion handling, vehicle platooning, and closing perception blind spots in complex traffic flows. These tasks are spatially dynamic: density and topology can change within hundreds of meters, and decision latency is constrained by vehicle speeds and safety margins. Fixed roadside units are effective for stable corridors but lack the agility to respond to short-term disruptions or localized coverage gaps. UAVs complement infrastructure by enabling on-demand aerial coverage and flexible network topology adaptation~\cite{zeng2016uav,mozaffari2016coverage}.

In a practical traffic management pipeline, UAV placement decisions would be triggered by short-term demand (e.g., queue buildup, incident reports, or V2X relay gaps) and updated on sub-second to minute scales depending on operational constraints. For example, a UAV tasked with providing temporary coverage for a congestion wave may need to reposition within tens of seconds, while the placement decision itself must be computed within a fraction of a second to allow integration with scheduling, routing, and safety checks. These system-level requirements motivate a deployment policy that is both accurate (to avoid coverage gaps) and fast (to enable real-time use).

Deploying UAVs in a motorway environment is not a single-objective placement problem. Operators must trade off coverage, link quality, and the number of UAVs while enforcing airspace and safety constraints. In our setting, a decision is required every 0.6~s (15 frames at 25~Hz in highD), and deployment decisions should remain within a typical ITS control loop (tens to hundreds of milliseconds). This makes high-quality optimization desirable but computationally challenging. Classical evolutionary solvers such as NSGA-II can approximate Pareto fronts, yet their per-scene runtime is often too slow for online use. Heuristic methods are fast but may underperform in critical regimes or fail to expose the full coverage--SNR--cost trade-off that deployment operators need.

\begin{figure}[t]
\centering
\includegraphics[width=\columnwidth]{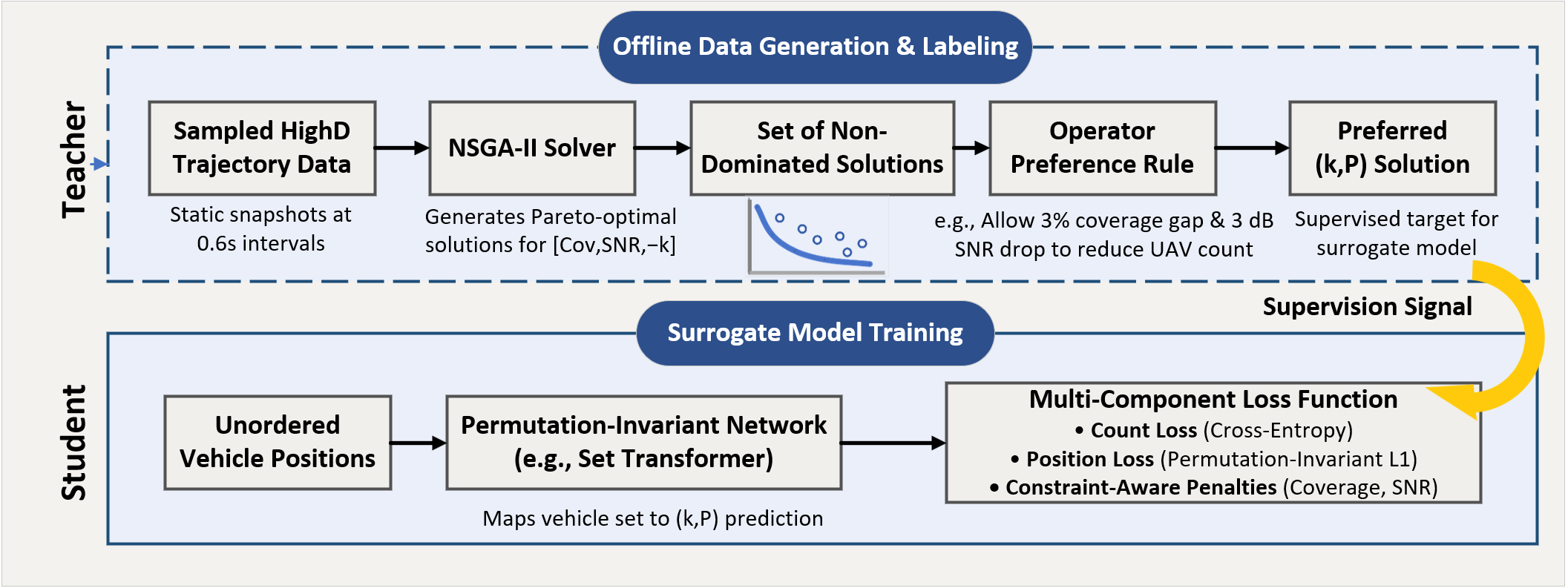}
\caption{Problem-to-deployment chain: motorway scenes are mapped to UAV decisions under multi-objective targets and constraints, then evaluated for real-time ITS deployment.}
\label{fig:problem_chain}
\end{figure}

This paper investigates whether near-NSGA-II multi-objective placement quality can be achieved with millisecond-scale inference suitable for motorway ITS, and whether this yields a better success--latency trade-off than heuristic baselines under a shared UAV budget. Fig.~\ref{fig:problem_chain} summarizes the end-to-end chain from traffic scene to deployment decision and evaluation. We formalize the task as a multi-objective discrete-continuous optimization problem with explicit bounds and separation constraints, then learn a surrogate predictor that maps vehicle point sets to both UAV count and positions. The surrogate is trained on NSGA-II-generated labels with a preference rule that reflects operational thresholds, and it incorporates permutation-aware learning to handle unordered vehicle sets.

A key challenge for ITS deployment is ensuring that the learned surrogate respects system constraints and is robust to data shifts across recordings. Real traffic data exhibit heterogeneous vehicle densities and lane usage patterns, which can lead to varying spatial extents and non-uniform cluster structures. As a result, naive sequence models may be brittle to input permutations or overfit to recording-specific statistics. We address these risks with permutation-invariant modeling, preference-aware label selection, and explicit constraint projection at inference time.

Compared with UAV-assisted ITS studies that prioritize trajectory or resource optimization without real-traffic generalization tests, we focus on static deployment decisions derived from motorway recordings and emphasize recording-level splits to reflect deployment shifts. Compared with learning-to-optimize studies that target fixed output dimensionality or synthetic inputs, we jointly predict the discrete UAV count and continuous placements from unordered vehicle point sets and evaluate multi-objective quality with Pareto metrics and deployment success curves. This framing connects the method to practical ITS decision loops while providing a reproducible benchmark and an evaluation protocol that can be audited by reviewers.

Our main model is a permutation-invariant Set Transformer, complemented by a sequence Transformer and other baselines to characterize inductive biases for unordered traffic scenes. We augment training with constraint-aware losses for coverage and SNR, and apply post-processing to enforce safety constraints. Evaluation follows recording-level splits to avoid leakage across traffic recordings and reports bootstrap confidence intervals, Pareto quality metrics, success-rate curves, and runtime benchmarks on both GPU and CPU. We further analyze robustness to sensor noise, scenario subgroups, and failure modes, and we provide deployment guidelines that link model choice to traffic conditions and latency constraints. The main contributions are:
\begin{itemize}
    \item \textbf{ITS-focused multi-objective benchmark:} a reproducible highD-derived dataset with recording-level splits, explicit constraints, and NSGA-II labels to evaluate UAV deployment under realistic motorway traffic.
    \item \textbf{Preference-guided surrogate learning:} a permutation-aware surrogate framework that predicts both UAV count and positions, with constraint-aware losses and post-processing for safe deployment.
    \item \textbf{Theoretical characterization:} properties of permutation-invariant matching, smooth objective surrogates, and exact-penalty feasibility that justify stable deployment-oriented training.
    \item \textbf{Deployment-oriented evaluation:} a multi-objective, statistically grounded protocol with success-rate curves, Pareto metrics, runtime benchmarks, and robustness analyses that expose system-level trade-offs; under a shared budget, the set-based surrogate achieves higher joint success than greedy/k-means.
\end{itemize}

The rest of the paper is organized as follows. Section~\ref{sec:related_work} reviews related work. Section~\ref{sec:problem_formulation} formalizes the placement problem and system assumptions. Section~\ref{sec:method} presents the algorithm design and learning framework. Section~\ref{sec:experiments_results} reports the dataset and experimental setup along with results, ablations, and discussion. Section~\ref{sec:conclusion} concludes the paper.

%% file: sections/related_work.tex
\section{Related Work}
\label{sec:related_work}
\subsection{UAV-Enabled ITS and Air--Ground Deployment}
UAV-assisted coverage has been studied in wireless communications and vehicular networks, where aerial base stations can fill coverage gaps or augment roadside infrastructure. Foundational studies analyze air-to-ground path loss and coverage geometry, including altitude selection for maximizing coverage~\cite{alhourani2014lap} and early UAV small-cell deployment analysis~\cite{mozaffari2015drone}. Surveys and system studies summarize opportunities and challenges of UAV communications in dynamic environments~\cite{zeng2016uav}. Recent ITS-focused works address UAV-assisted content delivery, V2X caching, and platoon-aware scheduling, often coupling trajectory planning with resource management~\cite{DBLP:journals/tits/Al-HiloSASE21,ZhangLuChengWangYang2021V2XDissemination,LiuZhouTianShengDuanQuLeung2022PlatoonMEC}. Additional T-ITS studies explore UAV-assisted RSUs and IoV services under practical constraints~\cite{AndreouMavromoustakisBatallaMarkakisMastorakis2023Voronoi,LiuLaiLinLeung2022MultiUAVIoV,QuBaiDaiLiuSun2025AoI}.
Complementary T-ITS work studies UAV-enabled data collection and wireless-powered two-way communication services, as well as RIS-UAV relaying for throughput enhancement~\cite{LiTanLiuVijayakumarKumarAlazab2021UAVSpeed,ParkHeoLeeLee2024TwoWay,LiuYuLiDurrani2022RISThroughput}. Recent T-ITS studies also investigate URLLC-aware proactive placement, traffic offloading in air--ground integrated networks, and weighted coverage deployment in UAV-enabled sensor systems~\cite{LiuWickramasingheSuraweeraBennisDebbah2024URLLC,FanJiangChenZhangWu2022Offloading,ZhuZhou2023WeightedCoverage}.

In the ITS context, deployment decisions must also align with transportation constraints such as lane-level coverage, incident response zones, and coordinated V2X scheduling. These system-level considerations motivate the use of real traffic data and evaluation protocols that respect cross-recording generalization, which are often absent in wireless-centric deployment studies. Security and authentication layers are also relevant to UAV-assisted vehicular systems~\cite{MiaoWangNingShankarMapleRodrigues2024AuthIoV,ChoiKwonSon2025PUF}, but are typically studied independently of placement decisions. TVT studies further cover QoE/cost-aware resource management, anti-jamming positioning, V2V-assisted 3D deployment, platooning control under uncertain links, blockchain-based edge delivery, temporary UAV-assisted VEC, and network-level modeling or interference-aware trajectory planning~\cite{FerrettiMignardiMariniVerdoneBuratti2024QoE,WangLiuLiuHanThompson2021AntiJamming,ZhangHeFengHan2024V2VAssistedUAV,DuanZhaoTianZhouMaZhang2024Platooning,XuJinSuLiWangFangWu2025BlockchainEdge,YangLiuLiLiXieXie2022TempUAVVEC,KhabbazAntounAssi2019Modeling,LiuZheng2024InterferenceAware}.

\subsection{Multi-Objective UAV Placement and Evolutionary Optimization}
Multi-objective formulations are common when coverage, SNR, energy, and UAV count must be jointly considered. NSGA-II remains a standard tool for Pareto front approximation~\cite{deb2002nsga2}, and recent wireless optimization studies extend multi-objective formulations to UAV deployment with energy and QoS constraints~\cite{zhu2024multi}. In vehicular networks, UAV trajectory and resource management is often optimized for mmWave links, beam tracking, or spectrum sharing~\cite{LiNiuWuAiHeWangChen2023mmWave,DBLP:journals/tvt/JangSKK23,QiSongGuoJamalipour2022SpectrumSharing}. Joint trajectory and radio resource optimization is also common in TVT studies~\cite{DBLP:journals/tvt/SpampinatoFBM25,ZhuLiuWangLiuChen2024Doppler}, reflecting the coupling between mobility and link quality. While these approaches can yield high-quality solutions, they require iterative optimization per scene, making them less suitable for low-latency ITS decision loops.

From a deployment perspective, multi-objective methods are valuable because they expose trade-offs that can be aligned with operator preferences (e.g., fewer UAVs vs. higher SNR). However, without a fast inference surrogate, these trade-offs cannot be exploited in online settings where hundreds of scenes may need to be evaluated per minute.

\subsection{Learning to Optimize on Unordered Inputs}
Learning-to-optimize replaces an expensive solver with a fast predictor trained on optimal or near-optimal labels~\cite{andrychowicz2016learning}. For unordered spatial inputs, permutation-invariant architectures such as DeepSets~\cite{zaheer2017deepsets} and Set Transformers~\cite{lee2019settransformer} provide principled aggregation, while point-specific attention models~\cite{zhao2021pointtransformer} and generic attention architectures~\cite{vaswani2017attention,jaegle2021perceiver} offer richer interactions. We follow this line in a traffic context: we learn a surrogate that predicts both the number and locations of UAVs from vehicle point sets, and we assess whether such surrogates can preserve Pareto quality and meet ITS runtime requirements.

Unlike many learning-to-optimize setups that assume fixed output dimensionality, UAV placement requires predicting both a discrete count and a set of continuous positions. This creates a permutation ambiguity in the regression target, motivating permutation-invariant losses or matching strategies. We address this ambiguity while maintaining the ability to evaluate multi-objective quality against an NSGA-II reference.

Overall, existing UAV deployment studies seldom evaluate on real traffic data with recording-level splits, and they rarely quantify multi-objective trade-offs using Pareto quality metrics or success-rate curves. Our work aims to close these gaps by combining real motorway data, preference-guided labeling, permutation-aware learning, and deployment-oriented evaluation that explicitly exposes count bias and runtime suitability.

\begin{table*}[t]
\centering
\caption{Representative ITS/UAV studies and their primary focus relative to our placement setting.}
\label{tab:prior_compare}
\small
\setlength{\tabcolsep}{4pt}
\begin{tabular}{lp{5.2cm}p{6.2cm}}
\toprule
Work & Primary focus (from title/abstract) & Relation to our setting \\
\midrule
Al-Hilo~\cite{DBLP:journals/tits/Al-HiloSASE21} & Joint UAV trajectory planning and cache management for ITS content delivery & We focus on static placement from traffic snapshots with multi-objective coverage/SNR/count; caching and trajectory planning are not modeled. \\
Liu~\cite{LiuZhouTianShengDuanQuLeung2022PlatoonMEC} & UAV-assisted MEC scheduling for platooning vehicles & Our setting replaces MEC scheduling with placement and emphasizes Pareto quality and deployment latency. \\
Andreou~\cite{AndreouMavromoustakisBatallaMarkakisMastorakis2023Voronoi} & Voronoi-based UAV-assisted RSU placement & We learn placement from real trajectories and evaluate multi-objective trade-offs and runtime, beyond geometric placement rules. \\
Liu~\cite{LiuLaiLinLeung2022MultiUAVIoV} & Joint communication and trajectory optimization for multi-UAV IoV & Our work targets instantaneous placement with count prediction instead of trajectory optimization. \\
Jang~\cite{DBLP:journals/tvt/JangSKK23} & Beam tracking for vehicular communications via UAV-assisted cellular networks & Physical-layer beam tracking is central, whereas placement and count prediction are central in our study. \\
Spampinato~\cite{DBLP:journals/tvt/SpampinatoFBM25} & Joint trajectory design and radio resource management & We isolate placement with multi-objective evaluation and real-traffic generalization rather than trajectory/resource co-optimization. \\
\bottomrule
\end{tabular}
\end{table*}

Table~\ref{tab:prior_compare} contrasts representative ITS/UAV studies with our placement setting. Most prior works emphasize trajectory/resource control or link-layer objectives, whereas we focus on instantaneous placement with count prediction, Pareto-quality evaluation, and real-traffic generalization. This positioning motivates our emphasis on recording-level splits, budgeted deployment policies, and runtime-aware evaluation.

%% file: sections/problem_formulation.tex
\section{Problem Formulation}
\label{sec:problem_formulation}
Consider a motorway scene with a set of vehicles $V = \{v_i \in \mathbb{R}^2\}_{i=1}^{N}$ in the ground plane. The task is to place $k$ UAVs with horizontal positions $P = \{p_j \in \mathbb{R}^2\}_{j=1}^{k}$, where $1 \le k \le K$ and $K=3$ in this work. Each UAV operates at a fixed altitude $h$ and must lie within scene bounds $(x_{\min}, x_{\max}, y_{\min}, y_{\max})$. A minimum separation constraint between active UAVs is enforced during post-processing to avoid degenerate colocations. The upper bound $K=3$ reflects a practical cap on simultaneous UAVs for a single motorway segment given coordination and cost constraints.

\subsection{System Model and Deployment Assumptions}
Each scenario is a bounded motorway segment derived from highD. The segment length is $403$~m on average (median $414$~m), and its width is $31$~m (median $31.5$~m), covering multi-lane traffic with consistent longitudinal flow. We assume a fixed UAV altitude of $120$~m, which aligns with typical low-altitude operational constraints and simplifies coordination with regulatory height limits. The communication model corresponds to downlink broadcast from UAVs to vehicles; coverage radius $R$ represents a service footprint for a target SNR threshold, while mean SNR captures link quality for continuous monitoring or V2X relay. The spatial bounds and minimum separation constraints correspond to geo-fencing and collision-avoidance requirements; they also limit co-channel interference by avoiding colocated aerial relays.

We interpret SNR thresholds as operator-defined service tiers in the robustness analysis: 43~dB for low-rate broadcast/monitoring, 45~dB for standard V2X messaging, and 47~dB for high-reliability demands. Table~\ref{tab:service_tiers} provides an illustrative mapping to target BLERs and representative MCS/throughput classes for a 10~MHz channel. This mapping is intended for interpretability and should be replaced by link-level MCS-versus-SNR curves of the target PHY; operators can then adjust policies by increasing $P_{\text{tx}}$, increasing the budgeted UAV count, or relaxing the threshold under adverse propagation.

\begin{table}[t]
\centering
\caption{Illustrative service-tier mapping for SNR thresholds (10~MHz channel).}
\label{tab:service_tiers}
\scriptsize
{\setlength{\tabcolsep}{3pt}
\begin{tabular}{lllll}
\toprule
Tier (SNR) & Service & BLER target & MCS & Spec. eff. (b/s/Hz) \\
\midrule
T1 (43~dB) & Broadcast/monitor & $\leq 10^{-2}$ & QPSK 1/2 & $\sim 1$~b/s/Hz \\
T2 (45~dB) & Standard V2X & $\leq 10^{-3}$ & 16QAM 1/2 & $\sim 2$~b/s/Hz \\
T3 (47~dB) & High rel. & $\leq 10^{-5}$ & 64QAM 1/2 & $\sim 3$~b/s/Hz \\
\bottomrule
\end{tabular}
}
\end{table}

\begin{table}[t]
\centering
\caption{System parameters with physical meaning and sources.}
\label{tab:params}
\scriptsize
\begin{tabular}{lll}
\toprule
Parameter & Value & Meaning / source \\
\midrule
UAV altitude $h$ & 120~m & Fixed low-altitude deployment \\
Coverage radius $R$ & 80~m & Service footprint for target SNR \\
Transmit power $P_{\text{tx}}$ & 30~dBm & Typical UAV TX budget \\
Noise floor $N_0$ & -95~dBm & 10~MHz, 9~dB NF \\
Path-loss exponent $\eta$ & 2.2 & Near-LoS urban macro \\
Reference loss $L_0$ & 32.4~dB (@1~m) & Free-space baseline \\
Min UAV separation & 15~m & Collision/safety buffer \\
\bottomrule
\end{tabular}
\end{table}
Unless otherwise noted, we use the parameter values in Table~\ref{tab:params}, including $h=120$~m, $R=80$~m, $P_{\text{tx}}=30$~dBm, $N_0=-95$~dBm (10~MHz bandwidth, 9~dB noise figure), and $\eta=2.2$.

\subsection{Coverage}
Let $R$ be the communication radius. A vehicle $v_i$ is covered if its horizontal distance to at least one UAV is within $R$:
\begin{equation}
\mathcal{C}(v_i, P) = \mathbb{I}\left( \min_{j \in \{1,\dots,k\}} \|v_i - p_j\|_2 \le R \right).
\end{equation}
The coverage ratio is
\begin{equation}
\text{Cov}(V, P) = \frac{1}{N} \sum_{i=1}^{N} \mathcal{C}(v_i, P).
\end{equation}

\subsection{Signal-to-Noise Ratio}
For each vehicle, the 3D distance to UAV $j$ is $d_{ij} = \sqrt{\|v_i - p_j\|_2^2 + h^2}$. With a path-loss model
\begin{equation}
L(d_{ij}) = L_0 + 10 \eta \log_{10}(d_{ij}),
\end{equation}
the received SNR in dB from the best UAV is
\begin{equation}
\text{SNR}_i = P_{\text{tx}} - L(d_{i*}) - N_0,
\end{equation}
where $d_{i*} = \min_j d_{ij}$, $P_{\text{tx}}$ is transmit power, and $N_0$ is the noise floor. We set $L_0=32.4$~dB at 1~m and $N_0=-95$~dBm, corresponding to a 10~MHz bandwidth with a 9~dB noise figure (Table~\ref{tab:params}). The mean SNR is
\begin{equation}
\text{SNR}(V, P) = \frac{1}{N} \sum_{i=1}^{N} \text{SNR}_i.
\end{equation}
We assume a noise-limited channel and do not model interference or shadowing; this corresponds to an orthogonal resource allocation across UAVs (e.g., time/frequency scheduling) and represents a best-case link model.

\subsection{Multi-Objective Formulation}
We seek UAV positions that maximize coverage and mean SNR while minimizing the number of UAVs:
\begin{equation}
\max_{P, k} \left[ \text{Cov}(V, P), \ \text{SNR}(V, P), \ -k \right],
\end{equation}
subject to $1 \le k \le K$, $p_j$ within bounds, and a minimum separation constraint between active UAVs:
\begin{equation}
\|p_a - p_b\|_2 \ge d_{\min}, \quad \forall a \ne b.
\end{equation}
The optimization is multi-objective and non-convex; we use NSGA-II to approximate Pareto-optimal solutions and select training labels.
The preference thresholds used to select labels are treated as operator policy parameters; alternative policies can be accommodated by re-labeling or by conditioning the surrogate on desired thresholds.

%% file: sections/method.tex
\section{Algorithm Design and Learning Framework}
\label{sec:method}
We propose a surrogate mapping from unordered vehicle positions to a joint decision: the UAV count class (1--3) and up to three UAV positions in the horizontal plane. The algorithmic contribution is a unified design that (i) predicts discrete counts and continuous placements from sets, (ii) removes label-order ambiguity through permutation-invariant matching, and (iii) aligns training with operational constraints via coverage/SNR surrogates and feasibility projection. Together, these components replace iterative multi-objective search with millisecond-scale inference while preserving deployable placements.

\begin{figure}[t]
\centering
\includegraphics[width=\columnwidth]{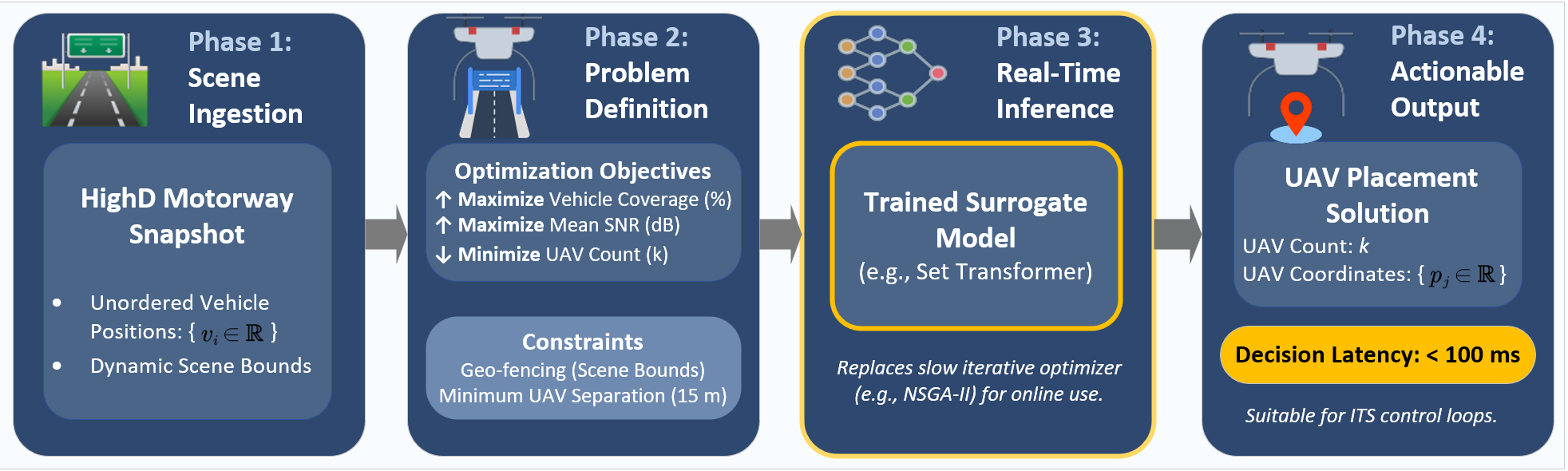}
\caption{End-to-end pipeline: highD scenarios are labeled by NSGA-II and used to train surrogate predictors for UAV count and placement.}
\label{fig:pipeline}
\end{figure}

\subsection{Surrogate Architecture}
Our main model is a Set Transformer~\cite{lee2019settransformer}, which is permutation-invariant by design and thus aligned with unordered vehicle inputs. Each vehicle $(x,y)$ is embedded into a $d=128$ token, then processed by two Induced Set Attention Blocks (ISAB) with 16 inducing points to capture long-range interactions at moderate complexity. A pooled multihead attention (PMA) module aggregates the set into a global scene representation. Two heads then predict the UAV count and UAV positions. This design preserves order invariance while allowing global context aggregation, which is useful for long motorway corridors with variable density.

\subsection{Permutation-Invariant Matching and Constraint-Aware Loss}
Because the UAV label set has no canonical ordering, we minimize a permutation-invariant $L_1$ loss that matches predicted UAV positions to the closest permutation of the ground-truth set, avoiding order ambiguity when $k>1$. We combine this with a cross-entropy loss for count classification so that the surrogate jointly predicts discrete and continuous decision variables. To align training with operational objectives, we augment the loss with smooth coverage and SNR surrogates and a separation penalty, nudging the model toward feasible placements without explicit iterative optimization.

\subsection{Post-Processing and Constraint Enforcement}
At inference, predicted UAV positions are optionally projected to satisfy scene bounds and a minimum separation constraint. The projection iteratively pushes UAV pairs apart when their distance is below 15~m and clamps positions to the scene bounds. We also evaluate a coverage-aware repair step that repositions UAVs toward uncovered vehicles while preserving separation, enabling a trade-off between constraint satisfaction and coverage. These steps emulate practical deployment policies where collision avoidance and geo-fencing are mandatory, and they provide explicit control over feasibility vs. performance.

\begin{algorithm}[t]
\caption{Deployment inference with constraint projection}
\label{alg:inference}
\begin{algorithmic}[1]
\STATE Input vehicle set $V$, bounds $(x_{\min}, x_{\max}, y_{\min}, y_{\max})$.
\STATE Predict count $\hat{k}$ and positions $\hat{P}$ with surrogate model.
\STATE Mask inactive UAV slots based on $\hat{k}$.
\STATE Project $\hat{P}$ to satisfy bounds and minimum separation.
\STATE Output feasible UAV placements $P^\star$.
\end{algorithmic}
\end{algorithm}

\subsection{Training Protocol}
\label{sec:training_protocol}
We split recordings into train/validation/test sets with a 60/20/20 ratio to avoid scene leakage. We create three independent split seeds (2025, 2026, 2027) and use the 2025 split for the main results reported here. Min-max scalers are fitted on the training split only and reused for validation and test data. All runs fix random seeds for NumPy and PyTorch. For sequence-based models, we apply random permutation augmentation of vehicle order during training to reduce ordering bias.

Each model jointly predicts UAV count and positions, aligning the surrogate with the decision variables in the optimization formulation. We use a cross-entropy loss for count classification and a permutation-invariant $L_1$ loss for positions that matches the predicted UAV set to the closest permutation of the ground-truth set. A weighted sum of count and position losses is optimized using Adam with early stopping (patience 20). Models are trained with a batch size of 64 and learning rate $10^{-3}$ unless otherwise noted by the architecture defaults.

We augment the training objective with constraint-aware penalties computed from predicted UAV locations and vehicle positions. Coverage is encouraged via a smooth coverage probability with sigmoid temperature $\tau=5$, SNR is encouraged via a softmin assignment with temperature $\tau=20$, and diversity can be enforced by penalizing UAV pairs closer than 15~m. We use a coverage target of 0.97 and minimum SNR of 45~dB (Table~\ref{tab:params}), with loss weights 5.0 (coverage) and 0.5 (SNR) for the main model. The diversity penalty is evaluated as an ablation.

\subsection{Theoretical Properties}
We summarize theoretical properties that justify the learning formulation and its alignment with the discrete deployment objectives. Let $d_{ij}=\lVert v_i-p_j\rVert_2$ denote horizontal distances, $\sigma(\cdot)$ be the logistic sigmoid, and define the smooth coverage and softmin distances used in training:
\begin{equation}
\begin{aligned}
\tilde{c}_i &= \max_j \sigma\left(\frac{R-d_{ij}}{\tau}\right), \\
\bar{d}_i &= \sum_j w_{ij} d_{ij}, \qquad
w_{ij}=\frac{\exp(-d_{ij}/\tau)}{\sum_{\ell}\exp(-d_{i\ell}/\tau)}.
\end{aligned}
\end{equation}
The mean soft coverage is $\tilde{\mathrm{Cov}} = \frac{1}{N}\sum_i \tilde{c}_i$ and the soft SNR is computed from $\bar{d}_i$ as in Section~\ref{sec:problem_formulation}.

\paragraph{Proposition 1 (Permutation-invariant matching).}
Let $\mathcal{L}_{\mathrm{match}}(P,G)=\min_{\pi\in S_k}\sum_{j=1}^k \lVert p_j-g_{\pi(j)}\rVert_1$. Then $\mathcal{L}_{\mathrm{match}}(P,G)$ is invariant to any permutation of $G$ (or $P$), and $\mathcal{L}_{\mathrm{match}}(P,G)=0$ iff the two sets are identical up to permutation.
\textit{Proof.} For any permutation $\sigma$, the feasible set $\{\pi\circ\sigma:\pi\in S_k\}$ equals $S_k$, so the minimum assignment cost is unchanged. If $P$ and $G$ match up to permutation, one permutation yields zero cost, and conversely zero cost implies elementwise equality under some permutation. \hfill $\square$

\paragraph{Proposition 2 (Consistency of smooth coverage).}
For each vehicle $i$, $\tilde{c}_i \to \mathbb{I}(\min_j d_{ij}\le R)$ pointwise as $\tau\to 0$, and therefore $\tilde{\mathrm{Cov}}\to \mathrm{Cov}$.
\textit{Proof.} Since $\sigma((R-d)/\tau)$ converges to the step function at $d=R$ as $\tau\to 0$, the maximum over $j$ converges to the indicator that at least one UAV lies within radius $R$. Averaging preserves convergence. \hfill $\square$

\paragraph{Proposition 3 (Softmin concentration).}
The softmin weights $w_{ij}$ concentrate on the nearest UAV as $\tau\to 0$, and $\bar{d}_i \to \min_j d_{ij}$.
\textit{Proof.} For $\tau\to 0$, the softmax over $-d_{ij}/\tau$ concentrates on the smallest distance index, yielding a one-hot distribution and thus $\bar{d}_i$ equal to the minimum distance. \hfill $\square$

\paragraph{Proposition 4 (Exact-penalty feasibility).}
Define the penalized loss
\begin{equation}
\begin{aligned}
\mathcal{L} &= \mathcal{L}_{\mathrm{match}} + \lambda_k \mathcal{L}_{\mathrm{count}} \\
&\quad + \lambda_c [C^\star-\tilde{\mathrm{Cov}}]_+ \\
&\quad + \lambda_s [S^\star-\tilde{\mathrm{SNR}}]_+ \\
&\quad + \lambda_d \sum_{a<b} [d_{\min}-\|p_a-p_b\|]_+^2.
\end{aligned}
\end{equation}
All penalty terms are nonnegative and vanish iff the corresponding constraint is satisfied. Hence any feasible solution is a minimizer of the penalty terms, and increasing $(\lambda_c,\lambda_s,\lambda_d)$ drives minimizers toward the feasible set whenever it is nonempty.
\textit{Proof.} Each hinge term is zero exactly when its constraint holds and positive otherwise, so feasible points incur no penalty. The standard exact-penalty argument applies by taking penalty weights sufficiently large. \hfill $\square$

\paragraph{Proposition 5 (Lipschitz stability).}
Let $P$ and $P'$ be two UAV sets with $\delta=\max_j \|p_j-p'_j\|_2$. Then
$|\tilde{c}_i(P)-\tilde{c}_i(P')| \le \delta/(4\tau)$ and
$|\tilde{\mathrm{Cov}}(P)-\tilde{\mathrm{Cov}}(P')| \le \delta/(4\tau)$.
Moreover, $|\bar{d}_i(P)-\bar{d}_i(P')| \le \delta$, and the induced soft SNR satisfies
$|\tilde{\mathrm{SNR}}(P)-\tilde{\mathrm{SNR}}(P')| \le L_s \delta$ for some constant $L_s$
depending on the path-loss parameters.
\textit{Proof.} $\sigma$ is $1/4$-Lipschitz, so $\sigma((R-d)/\tau)$ is $(4\tau)^{-1}$-Lipschitz in $d$.
The max and average preserve Lipschitz constants, and $d_{ij}$ is 1-Lipschitz in $p_j$.
For $\bar{d}_i$, all $d_{ij}$ shift by at most $\delta$, so the min and max shift by at most $\delta$,
and $\bar{d}_i$ lies between them. The SNR mapping is smooth on bounded distances, yielding a finite $L_s$.
\hfill $\square$

%% file: sections/experiments_results_discussion.tex
\section{Experiments, Results, and Discussion}
\label{sec:experiments_results}
\input{sections/dataset_method}
\input{sections/experiments}
\input{sections/results}
\input{sections/ablation}
\input{sections/discussion}

%% file: sections/dataset_method.tex
\subsection{Dataset and Labeling}
\label{sec:dataset_method}
\subsubsection{Dataset and Label Generation}
\label{sec:dataset_labeling}
We build a supervised dataset from the highD motorway recordings~\cite{krajewski2018highd}, captured by drones at 25~Hz over German highways with multiple lanes and unobstructed top-down views. We use 20 recordings (out of the full highD set) and sample frames every 15 time steps (0.6~s) up to 300 frames per recording to create static snapshots of vehicle positions. For each frame, we compute scene bounds from vehicle extents and add a small margin to allow UAV placement beyond the convex hull of vehicles. Vehicles within each frame are ordered by track ID (as provided by highD) to obtain a deterministic sequence for order-sensitive models; this ordering is randomized during training to mitigate permutation sensitivity (Section~\ref{sec:training_protocol}).

\begin{figure}[t]
\centering
\includegraphics[width=\columnwidth]{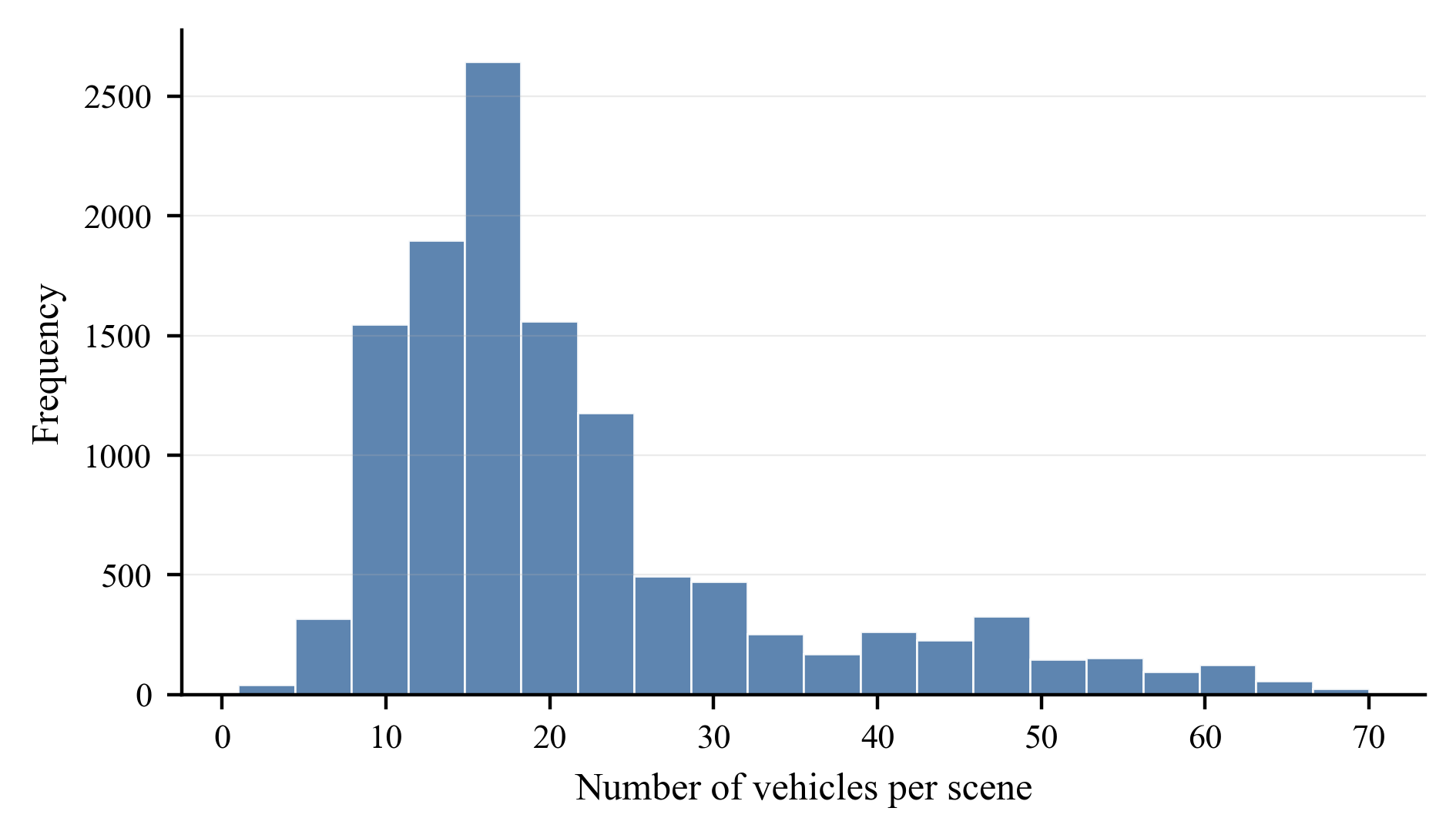}
\caption{Distribution of vehicle counts per scene.}
\label{fig:vehicle_count_hist}
\end{figure}

\paragraph{Dataset statistics.} The integrated dataset contains 12{,}000 scenarios spanning 20 recordings. Each sample includes up to 70 vehicle positions (mean 21.44, median 18, 95th percentile 50), a UAV count label in $\{1,2,3\}$, and up to three UAV positions. The distribution of vehicle counts is shown in Fig.~\ref{fig:vehicle_count_hist}. Per-recording vehicle counts range from 12.3 to 54.7 on average, reflecting diverse traffic densities. Recording-level splits are used to prevent leakage across training and test scenes; the split protocol is summarized in Section~\ref{sec:experiments}. Because frames within the same recording are temporally correlated, we report bootstrap confidence intervals at the recording level and include a sparser sampling ablation (30-frame stride) to validate robustness.

\paragraph{Recording statistics.} Table~\ref{tab:recording_stats} reports per-recording vehicle counts for the 20 highD recordings used in this study (600 sampled frames per recording).
\begin{table}[t]
\centering
\caption{Recording-level vehicle count statistics.}
\label{tab:recording_stats}
\scriptsize
{\setlength{\tabcolsep}{2pt}
\begin{tabular}{lcccc}
\toprule
Recording ID & Samples & Mean vehicles & Median vehicles & Max vehicles \\
\midrule
1 & 600 & 16.70 & 17 & 29 \\
2 & 600 & 14.50 & 14 & 32 \\
3 & 600 & 12.27 & 12 & 24 \\
4 & 600 & 14.76 & 15 & 25 \\
5 & 600 & 13.87 & 14 & 23 \\
6 & 600 & 16.17 & 16 & 25 \\
7 & 600 & 17.70 & 17 & 32 \\
8 & 600 & 16.84 & 16 & 31 \\
9 & 600 & 15.21 & 15 & 26 \\
10 & 600 & 18.63 & 19 & 37 \\
11 & 600 & 44.56 & 44 & 62 \\
12 & 600 & 54.68 & 56 & 70 \\
13 & 600 & 39.34 & 40 & 57 \\
15 & 600 & 17.14 & 17 & 35 \\
16 & 600 & 14.34 & 15 & 29 \\
18 & 600 & 13.71 & 14 & 24 \\
22 & 600 & 17.39 & 17 & 34 \\
23 & 600 & 18.56 & 19 & 32 \\
40 & 600 & 28.45 & 28 & 42 \\
58 & 600 & 23.97 & 23 & 40 \\
\bottomrule
\end{tabular}
}
\end{table}

\paragraph{NSGA-II label generation.} For each scenario, NSGA-II~\cite{deb2002nsga2} optimizes three objectives: coverage gap $(1-\text{Cov})$, UAV count $k$, and negative mean SNR $(-\text{SNR})$. We use a population size of 32, 60 generations, and allow $k \in [1,3]$. After obtaining the Pareto front, a preference rule selects a single label by (i) allowing a $3\%$ coverage gap from the best solution, (ii) allowing a $3$~dB drop in mean SNR within that subset, and (iii) preferring fewer UAVs. These thresholds align with our deployment targets of 97\% coverage and 45~dB mean SNR (Table~\ref{tab:params}) and reflect a typical operator policy: allow small quality drops to reduce the number of UAVs when the coverage/SNR difference is marginal. A sensitivity study over $(1\%,2\%,3\%,5\%)$ coverage tolerances and $(1,2,3,5)$~dB SNR tolerances shows that the selected $k$ distribution and mean objectives vary only slightly, indicating that the dataset is not overly sensitive to small preference changes. NSGA-II uses crossover rate 0.9, mutation rate 0.2, and a position mutation standard deviation of 15~m. NSGA-II labeling cost averages 0.41~s per scenario on our CPU (5 samples: 2.09~s), yielding a total labeling time of roughly 1.4~hours for 12{,}000 scenarios; this process is embarrassingly parallel.

UAV positions in the dataset are stored in fixed slots, but the label set has no canonical ordering. We therefore treat the labels as unordered and use permutation-invariant matching losses during training to eliminate label-order ambiguity.

\begin{algorithm}[t]
\caption{Label generation and surrogate training}
\label{alg:pipeline}
\begin{algorithmic}[1]
\STATE Sample frames from each recording and build vehicle sets $V$.
\FOR{each scenario}
    \STATE Run NSGA-II to obtain a Pareto front.
    \STATE Select a preferred solution under coverage/SNR tolerances.
    \STATE Store UAV count and positions as labels.
\ENDFOR
\STATE Train a surrogate predictor on $\{(V, \text{labels})\}$.
\end{algorithmic}
\end{algorithm}

\paragraph{Preference sensitivity.} Across coverage tolerances $\{1\%,2\%,3\%,5\%\}$ and SNR tolerances $\{1,2,3,5\}$~dB, the selected label distribution remains stable with mean $k$ between 2.62 and 2.65. Mean coverage stays within [0.9975, 0.9987] and mean SNR within [46.39, 46.40]~dB, indicating mild sensitivity to preference thresholds.

\begin{table}[t]
\centering
\caption{Preference sensitivity summary (NSGA-II labels).}
\label{tab:pref_sensitivity}
\scriptsize
{\setlength{\tabcolsep}{3pt}
\begin{tabular}{lccc}
\toprule
Tolerance (Cov., SNR) & Mean $k$ & Coverage & Mean SNR (dB) \\
\midrule
1\%, 1~dB & 2.65 & 0.9987 & 46.398 \\
5\%, 5~dB & 2.63 & 0.9975 & 46.390 \\
\bottomrule
\end{tabular}
}
\end{table}

%% file: sections/experiments.tex
\subsection{Experimental Setup}
\label{sec:experiments}
\subsubsection{Dataset and Splits}
We use the highD-derived dataset described in Section~\ref{sec:dataset_labeling}, containing 12{,}000 scenarios across 20 recordings. Recording-level splits are used to avoid scene leakage, with a 60/20/20 train/validation/test ratio. We create three independent splits (seeds 2025, 2026, 2027); the main paper reports the 2025 split with bootstrap confidence intervals, and additional split results are provided in the released code. Frames are sampled every 15 time steps (up to 300 per recording); a 30-frame stride ablation confirms that the main trends are robust to sparser sampling.

\subsubsection{Models, Baselines, and Training}
We compare against a greedy coverage heuristic, a k-means placement heuristic, and an oracle using the NSGA-II label solution. The learned models include a Transformer, a 1D CNN with Transformer head, a U-Net on occupancy grids, DeepSets, and a Set Transformer baseline.

All models are trained with Adam (learning rate $10^{-3}$, batch size 64) and early stopping on the validation split. The Transformer uses a 3-layer encoder with 8 heads; the 1D CNN uses residual blocks with attention and a 3-layer Transformer head; the U-Net operates on $128 \times 128$ occupancy grids; and DeepSets uses mean/max pooling. K-means uses 20 iterations, and the greedy baseline selects UAVs by incremental coverage gain. For post-processing, we apply 10 projection iterations and a minimum separation of 15~m. Each reported model is trained once per split with the same data and optimization budget.

\subsubsection{Metrics and Evaluation Protocol}
We report UAV count accuracy, mean coverage ratio, and mean SNR on the test recordings. Confidence intervals are computed with 500 bootstrap resamples over recordings. We additionally report success rates for coverage $\geq 0.95$ and mean SNR $\geq 45$~dB, and track the predicted count distribution to quantify over/under-deployment. To expose operational trade-offs, we evaluate success-rate curves over coverage and SNR thresholds (Section~\ref{sec:results}). Predicted UAV positions are optionally post-processed to enforce bounds and a minimum separation of 15~m.

For multi-objective quality, we compute generational distance (GD), inverted generational distance (IGD), and dominated hypervolume (HV) in a normalized objective space. Pareto fronts are approximated with NSGA-II for a subset of 100 test scenarios. Objectives are normalized by coverage gap in $[0,1]$, UAV count divided by $k_{\mathrm{ref}}=4$, and negative SNR divided by 120, with a reference point at $(1,1,1)$. We further report paired Wilcoxon tests for GD/IGD/HV against the Transformer baseline.

All learned models receive the same vehicle positions as inputs and are evaluated with identical normalization and post-processing policies. Models are evaluated under two masking policies: using the ground-truth count (mask-by true) and using the predicted count (mask-by pred). For heuristics, we use the ground-truth UAV count since these methods do not predict $k$. Hyperparameters for all baselines are fixed across splits; no per-recording tuning is performed. We report recording-level bootstrap confidence intervals to account for within-recording temporal correlation.

\subsubsection{Hardware, Software, and Runtime Protocol}
All experiments are run on a Windows 11 workstation with an Intel Core i7-13790F CPU (24 logical processors) and an NVIDIA RTX 4090 GPU. The software stack includes Python 3.12.7, PyTorch 2.7.1, NumPy 1.26.4, and Pandas 2.2.2.

Inference runtime is measured on 200 test samples for GPU and 50 test samples for CPU, while NSGA-II timing is measured on 10 (GPU) or 5 (CPU) samples due to its cost. We report end-to-end latency (including preprocessing and post-processing) for batch sizes 1 and 64 on GPU, and batch size 1 on CPU, plus throughput in samples per second. Section~\ref{sec:results} summarizes the batch size 1 results.

%% file: sections/results.tex
\subsection{Results and Analysis}
\label{sec:results}
Table~\ref{tab:main_results} reports cross-recording results on the seed-2025 split under the budget-matched count policy (Policy~B, mean $k=2.74$). The Set Transformer achieves the strongest coverage--SNR trade-off among learned models and yields the highest joint success rate at the shared budget (0.791). Greedy and k-means remain competitive on coverage and SNR but are consistently lower in joint success, while the facility-location baseline (k-medoids) trails k-means/greedy under the shared budget. The Transformer benefits from constraint-aware training but remains less accurate in count prediction (0.747 vs. 0.950 for the Set Transformer). Bootstrap confidence intervals are computed over four test recordings.

\begin{table*}[t]
\centering
\caption{Cross-recording results under Policy~B (budget-matched $k$, mean $k=2.74$). Success uses Cov$\geq$0.95 and SNR$\geq$45~dB.}
\label{tab:main_results}
\small
\begin{tabular}{lccc}
\toprule
Model & Coverage & Mean SNR (dB) & Success rate \\
\midrule
Transformer & 0.928 [0.901, 0.956] & 46.132 [46.044, 46.219] & 0.625 [0.521, 0.730] \\
Set Transformer & 0.957 [0.936, 0.977] & 46.252 [46.179, 46.325] & 0.791 [0.708, 0.873] \\
1D CNN + Transformer & 0.527 [0.524, 0.531] & 44.877 [44.862, 44.889] & 0.003 [0.000, 0.006] \\
Picture CNN (U-Net) & 0.475 [0.467, 0.483] & 44.497 [44.450, 44.533] & 0.003 [0.001, 0.006] \\
DeepSets & 0.583 [0.564, 0.597] & 45.091 [45.027, 45.152] & 0.013 [0.005, 0.021] \\
Greedy & 0.955 [0.942, 0.967] & 46.208 [46.175, 46.239] & 0.644 [0.573, 0.715] \\
K-means & 0.964 [0.953, 0.975] & 46.374 [46.342, 46.405] & 0.760 [0.703, 0.819] \\
K-medoids & 0.941 [0.923, 0.959] & 46.339 [46.301, 46.373] & 0.588 [0.516, 0.683] \\
\midrule
\multicolumn{4}{l}{\emph{Upper bound (not budgeted)}} \\
Oracle (preferred NSGA-II label) & 0.967 [0.953, 0.981] & 46.293 [46.231, 46.354] & 0.881 [0.828, 0.933] \\
\bottomrule
\end{tabular}
\end{table*}

Across bootstrap resamples, the confidence intervals remain narrow for coverage and SNR, indicating stable generalization. Coverage shows modest variance for the Transformer but remains consistently above the CNN and DeepSets baselines. Using success thresholds of coverage $\geq 0.95$ and SNR $\geq 45$~dB, the Set Transformer achieves a success rate of 0.791, outperforming greedy (0.644), k-means (0.760), and k-medoids (0.588), while remaining below the oracle upper bound (0.881).

\begin{table}[t]
\centering
\caption{Runtime benchmark on GPU (batch size 1, end-to-end ms, speedup vs. NSGA-II).}
\label{tab:runtime}
\small
\begin{tabular}{lccc}
\toprule
Method & Params (M) & Time (ms) & Speedup \\
\midrule
Transformer & 4.08 & 1.20 & 3785$\times$ \\
Set Transformer & 0.54 & 2.44 & 1870$\times$ \\
1D CNN + Transformer & 8.32 & 4.59 & 993$\times$ \\
Picture CNN (U-Net) & 6.12 & 1.95 & 2342$\times$ \\
DeepSets & 0.09 & 0.80 & 5719$\times$ \\
Greedy & -- & 54.19 & 84$\times$ \\
K-means & -- & 105.88 & 43$\times$ \\
K-medoids & -- & 18.18 & 251$\times$ \\
NSGA-II & -- & 4556.1 & 1$\times$ \\
\bottomrule
\end{tabular}
\end{table}

\begin{table}[t]
\centering
\caption{Deployment success vs. latency under Policy~B (success: Cov$\geq$0.95 and SNR$\geq$45~dB).}
\label{tab:deploy_tradeoff}
\small
\begin{tabular}{lccc}
\toprule
Method & Success rate & GPU ms & CPU ms \\
\midrule
Set Transformer & 0.791 & 2.44 & 4.01 \\
Transformer & 0.625 & 1.20 & 13.45 \\
Greedy & 0.644 & 54.19 & 64.72 \\
K-means & 0.760 & 105.88 & 116.45 \\
K-medoids & 0.588 & 18.18 & 19.68 \\
NSGA-II (oracle) & 0.881 & 4556.1 & 4565.7 \\
\bottomrule
\end{tabular}
\end{table}

\begin{figure}[t]
\centering
\includegraphics[width=\columnwidth]{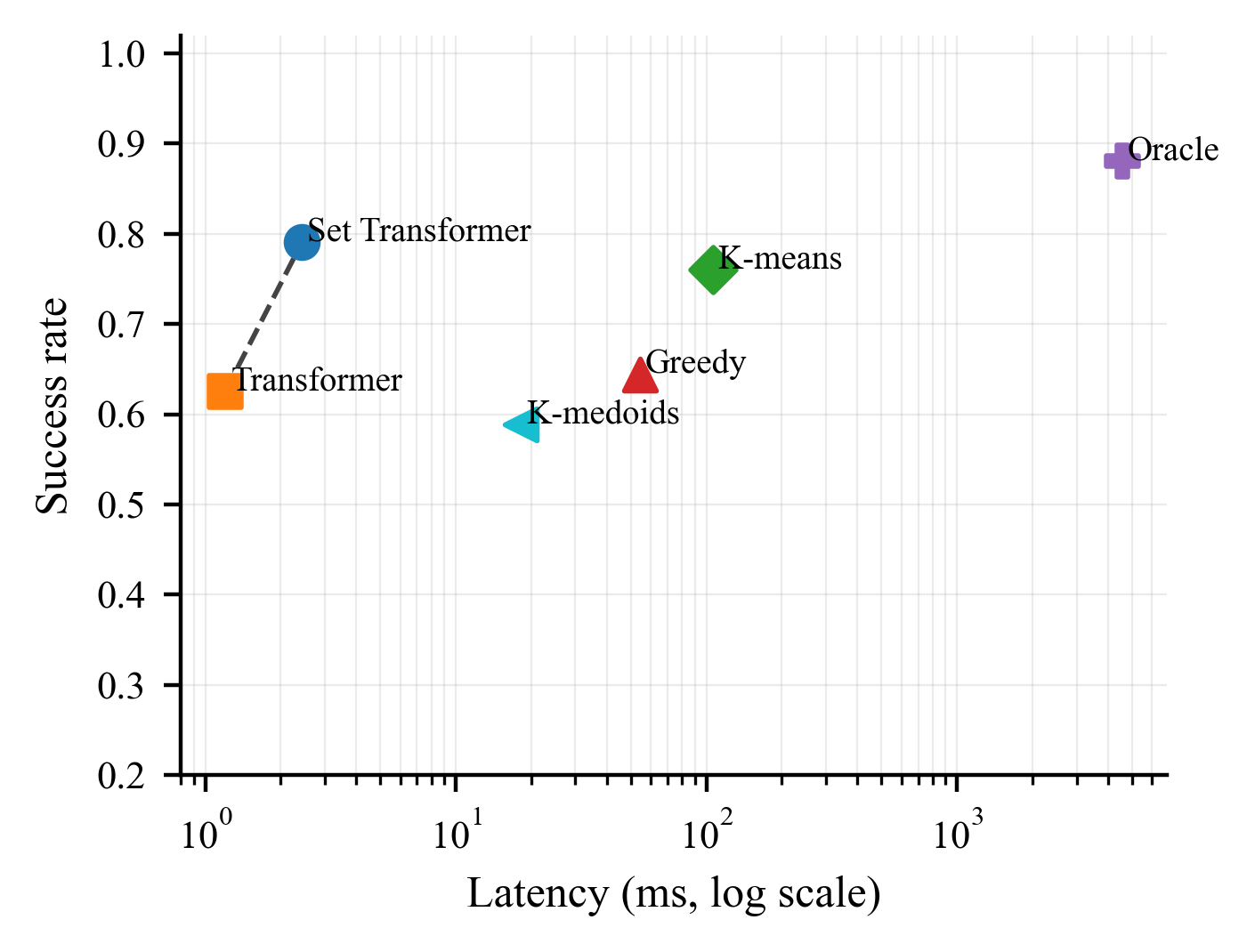}
\caption{Success--latency scatter under Policy~B (log-scale latency). The dashed line connects non-dominated points.}
\label{fig:success_latency}
\end{figure}

Fig.~\ref{fig:success_latency} summarizes the success--latency trade-off under Policy~B. The Set Transformer dominates greedy, k-means, and k-medoids by achieving higher joint success at substantially lower latency, while the oracle upper bound remains far slower.

Table~\ref{tab:pareto_metrics} reports Pareto quality metrics computed on 100 sampled test scenarios (mask-by true). Lower GD/IGD and higher HV indicate closer alignment to the NSGA-II front. Set Transformer attains the lowest GD among learned models, while the Transformer achieves competitive HV; paired Wilcoxon tests confirm statistically significant differences for most baselines.

\begin{table*}[t]
\centering
\caption{Pareto metrics on sampled test scenarios (mean with 95\% CI, mask-by true).}
\label{tab:pareto_metrics}
\small
\begin{tabular}{lccc}
\toprule
Model & GD $\downarrow$ & IGD $\downarrow$ & HV $\uparrow$ \\
\midrule
Transformer & 0.012 [0.004, 0.025] & 0.445 [0.422, 0.465] & 0.406 [0.384, 0.431] \\
Set Transformer & 0.007 [0.003, 0.012] & 0.462 [0.441, 0.481] & 0.436 [0.410, 0.471] \\
DeepSets & 0.046 [0.034, 0.057] & 0.379 [0.364, 0.393] & 0.292 [0.263, 0.328] \\
Greedy & 0.003 [0.002, 0.006] & 0.454 [0.435, 0.472] & 0.435 [0.408, 0.471] \\
K-means & 0.005 [0.001, 0.011] & 0.469 [0.449, 0.489] & 0.447 [0.418, 0.485] \\
Oracle (NSGA-II label) & 0.009 [0.004, 0.015] & 0.484 [0.468, 0.501] & 0.461 [0.429, 0.502] \\
\bottomrule
\end{tabular}
\end{table*}

Among learned models, Set Transformer attains the lowest GD (0.007) and the highest HV (0.436), indicating closer alignment to the NSGA-II front. DeepSets exhibits lower IGD but substantially lower HV, suggesting that its solutions are less competitive across the full objective range.

The runtime results highlight a clear trade-off between inference speed and coverage quality. DeepSets is the fastest learned model, while the Transformer and Set Transformer provide a strong balance between accuracy and speed and remain well within typical ITS real-time constraints (e.g., $<$100~ms per decision). On CPU, batch-1 end-to-end latency is 13.45~ms for the Transformer and 4.01~ms for the Set Transformer, still far below typical control-loop budgets.
Table~\ref{tab:deploy_tradeoff} summarizes deployment-level trade-offs using a joint success criterion (coverage $\geq 0.95$ and SNR $\geq 45$~dB). The Set Transformer yields the highest success rate among learned models while remaining three orders of magnitude faster than NSGA-II, whereas the Transformer offers lower success with the lowest GPU latency.

\subsubsection{Sensitivity Analyses}
\begin{figure}[t]
\centering
\includegraphics[width=\columnwidth]{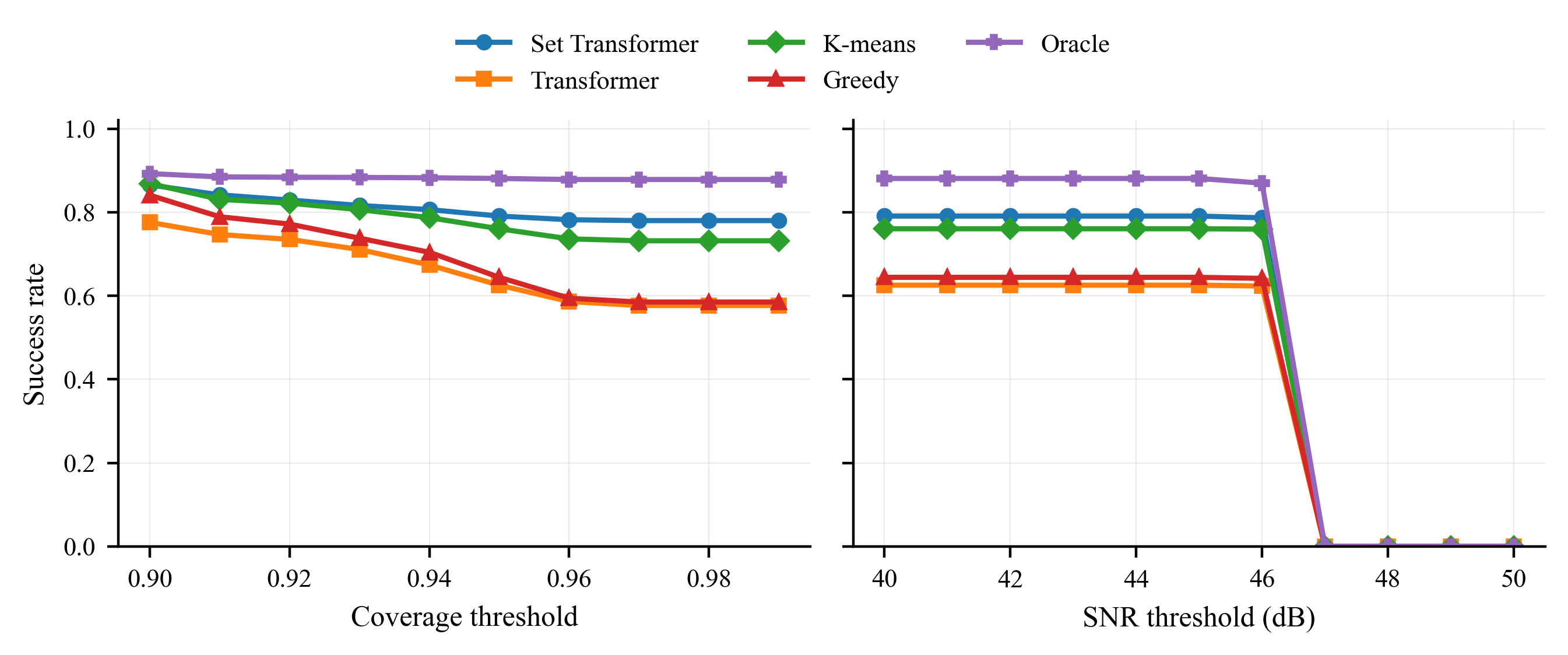}
\caption{Success-rate curves versus coverage and SNR thresholds under Policy~B (budget-matched $k$).}
\label{fig:success_curve}
\end{figure}

Fig.~\ref{fig:success_curve} shows success rates as the coverage and SNR thresholds vary under Policy~B. At a fixed SNR threshold of 45~dB, the Set Transformer maintains a success rate of 0.791 at 95\% coverage and 0.780 at 98\% coverage, while the Transformer achieves 0.625 and 0.577, respectively. K-means remains competitive at moderate thresholds (0.760 at 95\% coverage) but degrades more sharply as coverage thresholds increase. Success rates remain flat between 40 and 46~dB and collapse at 47~dB, indicating that the mean SNR distribution concentrates near 46~dB for all methods.

We further evaluate robustness to position noise to reflect localization errors in deployment.

\begin{figure}[t]
\centering
\includegraphics[width=\columnwidth]{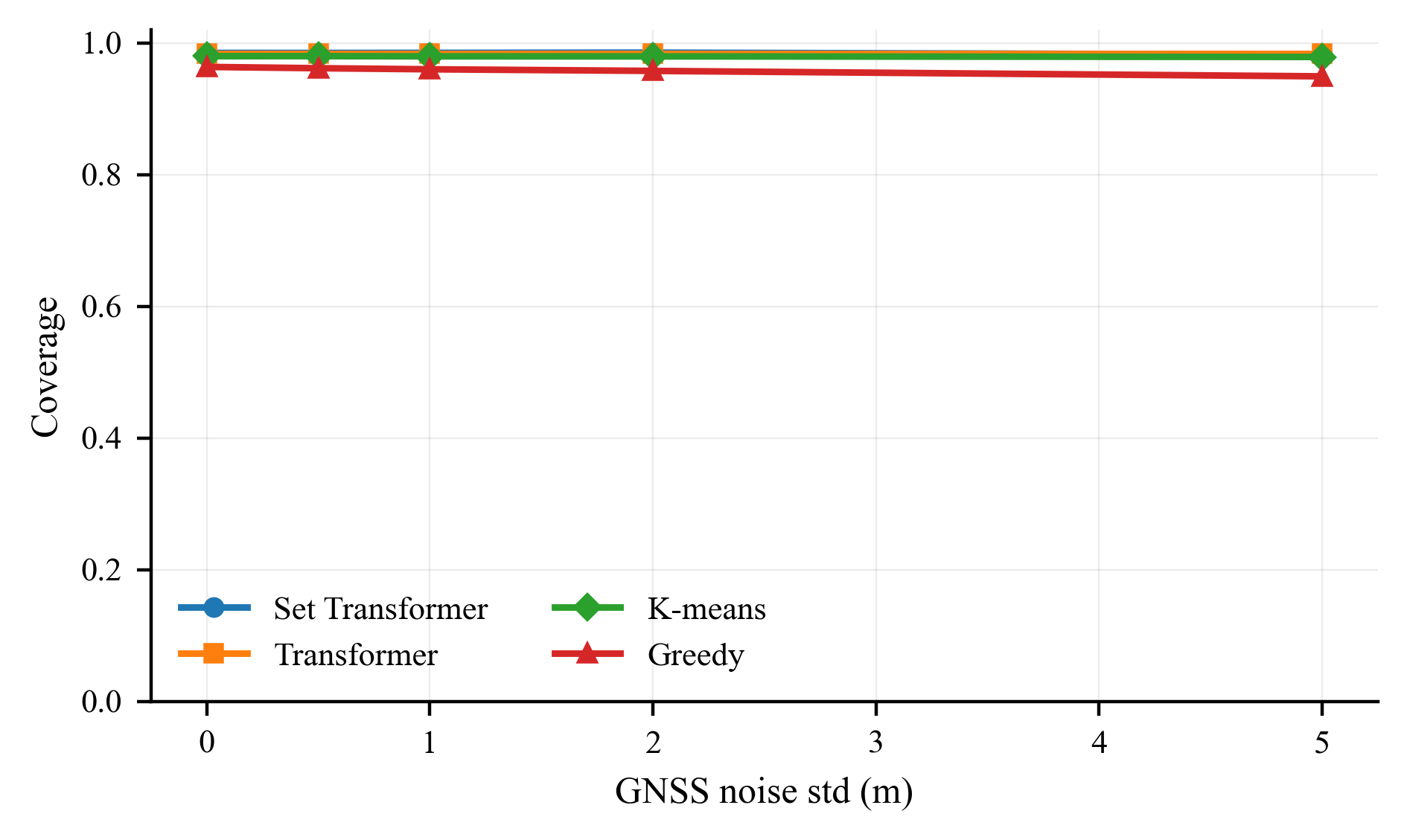}
\caption{Coverage under GNSS noise (mask-by pred for learned models, mask-by true for heuristics).}
\label{fig:noise_sensitivity}
\end{figure}

Fig.~\ref{fig:noise_sensitivity} reports sensitivity to Gaussian GNSS noise in vehicle positions. The Transformer shows negligible change in coverage and success up to 5~m noise (0.825 throughout), and the Set Transformer degrades mildly (success drops from 0.865 to 0.853 at 5~m). Greedy is more sensitive: success falls from 0.668 to 0.551, indicating that learned models are more robust to moderate localization noise.
These bounded degradations are consistent with the Lipschitz stability in Proposition~5. The permutation sensitivity results in Section~\ref{sec:ablation} empirically support Proposition~1.

\subsubsection{Failure Case and Subset Analysis}
\begin{figure}[t]
\centering
\includegraphics[width=0.8\columnwidth]{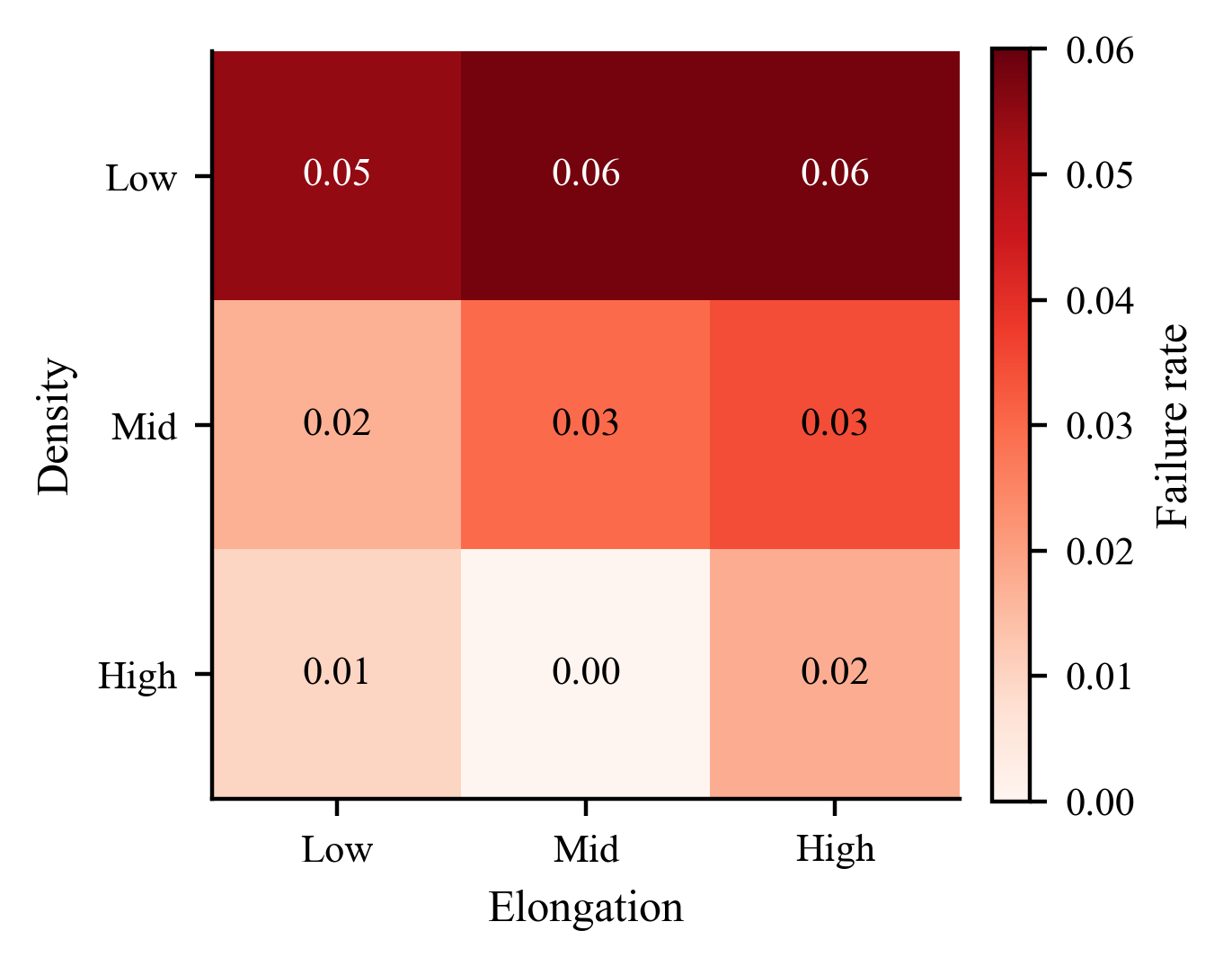}
\caption{Failure-rate heatmap by density and elongation (Transformer, mask-by pred).}
\label{fig:failure_heatmap}
\end{figure}

Table~\ref{tab:failure_clusters} summarizes failure rates across density and elongation clusters, where failure is coverage $<0.90$ or SNR $<45$~dB. Low-density and highly elongated scenes exhibit the highest failure rates (5--6\%). High-density scenes with moderate elongation have near-zero failure rates, and Fig.~\ref{fig:failure_heatmap} visualizes this trend.

\begin{table}[t]
\centering
\caption{Failure rates by density and elongation cluster (Transformer, mask-by pred).}
\label{tab:failure_clusters}
\scriptsize
{\setlength{\tabcolsep}{3pt}
\begin{tabular}{lccc}
\toprule
Cluster & Failure rate & Coverage & Mean SNR (dB) \\
\midrule
Low dens., low el. & 0.054 & 0.986 & 46.320 \\
Low dens., mid el. & 0.058 & 0.981 & 46.322 \\
Low dens., high el. & 0.058 & 0.979 & 46.273 \\
Mid dens., low el. & 0.017 & 0.988 & 46.327 \\
Mid dens., mid el. & 0.030 & 0.985 & 46.334 \\
Mid dens., high el. & 0.034 & 0.977 & 46.299 \\
High dens., low el. & 0.010 & 0.984 & 46.322 \\
High dens., mid el. & 0.000 & 0.987 & 46.339 \\
High dens., high el. & 0.017 & 0.984 & 46.322 \\
\bottomrule
\end{tabular}
}
\end{table}

\begin{table}[t]
\centering
\caption{Clustered success rates under Policy~B (budgeted $k$). Delta is Set Transformer minus the best heuristic.}
\label{tab:subset_success}
\scriptsize
\begin{tabular}{lccccc}
\toprule
Cluster & $N$ & Set Trans. & Greedy & K-means & Delta \\
\midrule
Low / Low & 294 & 0.673 & 0.469 & 0.605 & +0.068 \\
Low / Mid & 258 & 0.465 & 0.341 & 0.419 & +0.047 \\
Low / High & 240 & 0.417 & 0.300 & 0.367 & +0.050 \\
Mid / Low & 293 & 0.843 & 0.799 & 0.867 & -0.024 \\
Mid / Mid & 267 & 0.921 & 0.727 & 0.873 & +0.049 \\
Mid / High & 232 & 0.897 & 0.616 & 0.810 & +0.086 \\
High / Low & 205 & 0.888 & 0.932 & 0.985 & -0.098 \\
High / Mid & 267 & 0.993 & 0.858 & 0.970 & +0.022 \\
High / High & 344 & 0.965 & 0.747 & 0.916 & +0.049 \\
\bottomrule
\end{tabular}
\end{table}

To assess failure modes, we inspect low-coverage Transformer predictions on the test split. A representative case shows vehicles occupying a long, sparse corridor and the predicted UAVs clustering near one region, leaving distant vehicles uncovered. A cluster analysis confirms higher failure rates (5--6\%) in low-density, highly elongated scenes, while dense scenes exhibit near-zero failure rates.

Table~\ref{tab:subset_success} compares joint success across density/elongation clusters under the shared budget (Policy~B). The Set Transformer exceeds the best heuristic in 7 of 9 clusters, with the largest gains in elongated scenes (e.g., +0.086 in mid/high and +0.049 in high/high), while k-means remains strongest in compact high-density scenes (high/low). These cases highlight the need for stronger spatial diversity or explicit trajectory coupling when scene extents become highly elongated.

%% file: sections/ablation.tex
\subsection{Ablation Studies}
\label{sec:ablation}
We report four ablations covering training recipe, masking policy, permutation sensitivity, and post-processing. Table~\ref{tab:training_ablation} compares a baseline Transformer trained with count cross-entropy and $L_1$ regression against the proposed recipe with permutation-invariant matching, random-order augmentation, and constraint-aware losses. The proposed recipe improves coverage by 0.150 while keeping count accuracy unchanged.

\begin{table}[t]
\centering
\caption{Training-recipe ablation for the Transformer (mask-by true, mean with 95\% CI).}
\label{tab:training_ablation}
\scriptsize
{\setlength{\tabcolsep}{2pt}
\begin{tabular}{lccc}
\toprule
Model & Count Acc. & Coverage & Mean SNR (dB) \\
\midrule
Baseline (CE+$L_1$) & 0.747 [0.648,0.840] & 0.779 [0.762,0.796] & 45.872 [45.807,45.944] \\
Proposed recipe & 0.747 [0.648,0.840] & 0.929 [0.907,0.950] & 46.138 [46.063,46.211] \\
\bottomrule
\end{tabular}
}
\end{table}

Table~\ref{tab:mask_ablation} compares evaluation with ground-truth UAV counts (mask-by true) versus predicted counts (mask-by pred). The Transformer benefits most from mask-by pred because it overpredicts $k=3$ (mean 3.00, over-rate 25.3\%). This increases coverage but can deviate from the count minimization objective, so we report Pareto metrics under both count sources (Section~\ref{sec:results}).

\begin{table*}[t]
\centering
\caption{Mask-policy ablation using predicted-count masking (mean with 95\% CI).}
\label{tab:mask_ablation}
\scriptsize
{\setlength{\tabcolsep}{3pt}
\begin{tabular}{lcc}
\toprule
Model & Coverage & Mean SNR (dB) \\
\midrule
Transformer & 0.984 [0.982,0.985] & 46.318 [46.301,46.330] \\
Set Transformer & 0.985 [0.981,0.989] & 46.369 [46.354,46.387] \\
1D CNN + Transformer & 0.527 [0.524,0.530] & 44.874 [44.862,44.886] \\
Picture CNN (U-Net) & 0.475 [0.465,0.485] & 44.503 [44.446,44.553] \\
DeepSets & 0.585 [0.566,0.600] & 45.118 [45.044,45.193] \\
\bottomrule
\end{tabular}
}
\end{table*}

Permutation sensitivity is evaluated by averaging 10 random orderings per test sample. The average per-sample standard deviation across permutations is $1.7\times 10^{-8}$ for coverage and $6.9\times 10^{-6}$~dB for SNR, indicating near permutation invariance under the proposed training recipe.

\begin{figure}[t]
\centering
\includegraphics[width=\columnwidth]{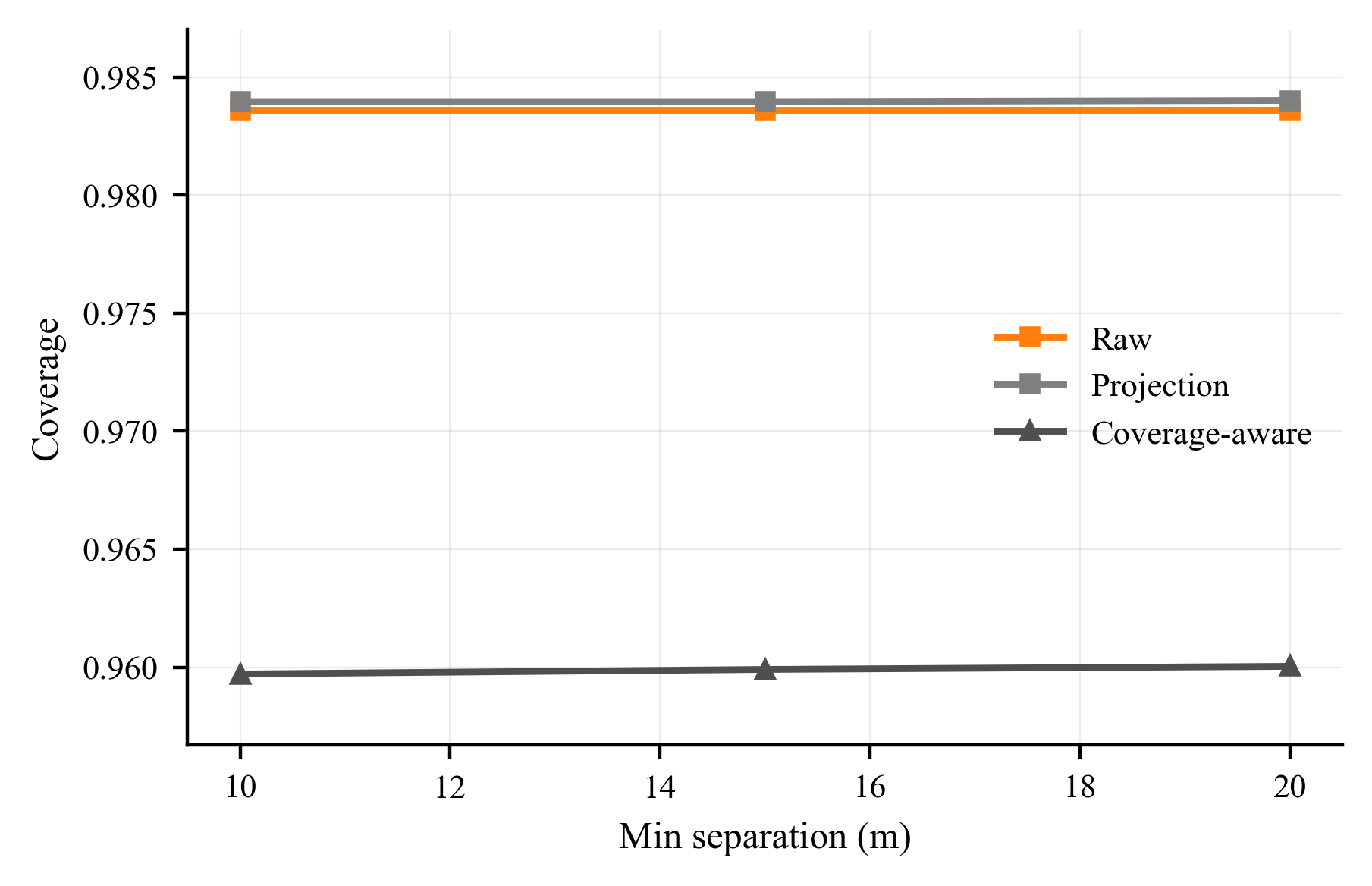}
\caption{Post-processing trade-off between coverage and constraint violations at different separation thresholds.}
\label{fig:postprocess_tradeoff}
\end{figure}

For post-processing, Table~\ref{tab:postprocess_ablation} summarizes the Transformer (mask-by pred) before and after projection. Simple projection removes out-of-bounds UAVs without reducing coverage. A coverage-aware repair step further reduces constraint violations at a cost of 2.37 points in coverage. Fig.~\ref{fig:postprocess_tradeoff} visualizes the coverage--constraint trade-off across separation thresholds (10/15/20~m).

\begin{table}[t]
\centering
\caption{Post-processing ablation for the Transformer (mask-by pred). OOB: out-of-bounds rate.}
\label{tab:postprocess_ablation}
\scriptsize
{\setlength{\tabcolsep}{2pt}
\begin{tabular}{lcccc}
\toprule
Setting & Coverage & SNR (dB) & OOB & Sep. viol. \\
\midrule
Raw & 0.984 & 46.318 & 0.0408 & 0.0000 \\
Projected & 0.984 & 46.327 & 0.0000 & 0.0013 \\
Cov.-aware repair & 0.960 & 46.264 & 0.0042 & 0.0004 \\
\bottomrule
\end{tabular}
}
\end{table}

%% file: sections/discussion.tex
\subsection{Discussion and Limitations}
\label{sec:discussion}
Our study relies on static snapshots and a simplified air-to-ground channel model. In practice, UAVs must coordinate over time with vehicle motion, energy constraints, and no-fly zones, and propagation is affected by shadowing, interference, and non-line-of-sight conditions. A sensitivity analysis over path-loss exponents $\eta \in \{2.0, 2.2, 2.6, 3.0\}$ shows that coverage remains stable but mean SNR drops from 50.5~dB to 29.5~dB, indicating that deployment performance is highly channel dependent and should be re-optimized when propagation conditions change.

The learned predictors inherit the preference encoded by the NSGA-II label selection (coverage and SNR tolerances). A tolerance sweep yields a stable label distribution (mean $k$ between 2.62 and 2.65), but the Transformer still overpredicts $k=3$ at inference, which boosts coverage under mask-by pred yet deviates from the count minimization objective. Addressing this bias may require preference-conditioned models, explicit count calibration, or hybrid inference that falls back to heuristics under low confidence. Permutation sensitivity is negligible after random-order augmentation, yet strongly elongated scenes remain failure modes, suggesting a need for stronger diversity constraints or global context.

Scalability beyond $K=3$ remains a challenge. A $K=5$ Transformer trained on an expanded 960-scenario subset reaches coverage 0.761, mean SNR 45.77~dB, and count accuracy 0.721, still below the $K=3$ regime, indicating that larger output sets require richer supervision, slot-based assignment, or permutation-invariant matching at higher capacity. This is a key direction for future work.

From a safety and regulatory standpoint, deployments must respect geo-fencing, altitude limits, and separation constraints. Our post-processing provides a lightweight feasibility projection, but practical systems should also handle no-fly zones, dynamic altitude corridors, and explicit collision-avoidance policies. We also find that learned models are robust to moderate GNSS noise (up to 5~m), while greedy heuristics degrade more noticeably; however, larger localization errors or delayed sensing could still lead to unsafe placements. A practical deployment should include uncertainty estimation and fallback policies (e.g., switching to a heuristic or short-horizon optimization when confidence is low), as well as monitoring for failure modes in elongated corridors.

\subsubsection{Practical Deployment Guidelines}
Our results suggest the following operational choices. The Set Transformer is preferred when coverage reliability is critical: it maintains the highest success rate across thresholds and is robust to moderate position noise. The Transformer offers a strong accuracy--speed balance and is competitive when compute budgets are tight but coverage thresholds are moderate. DeepSets is suitable for ultra-low latency when coverage is not the dominant objective. In sparse, elongated corridors, heuristic baselines remain strong; a hybrid policy can use heuristics for low-density scenes and learned models for dense or clustered traffic. These guidelines are derived from the success-rate curves, runtime benchmarks, and subset analyses in Section~\ref{sec:results}.

To support reproducibility, we will release the code and recording-level splits in a public repository (\href{https://github.com/highd-uav-its/highd-uav-its}{github.com/highd-uav-its}). The highD dataset is publicly available; we will provide the derived labels and train/val/test splits. Future work will incorporate temporal models, more realistic channel models, and additional datasets or real-world measurements to validate deployment robustness.

%% file: sections/conclusion.tex
\section{Conclusion}
\label{sec:conclusion}
We presented a surrogate-learning framework for multi-objective UAV placement in motorway ITS. Using NSGA-II labels from highD recordings, we trained multiple predictors and evaluated them under recording-level splits with confidence intervals and runtime benchmarks. The Set Transformer achieves the highest coverage among learned models, while the Transformer provides a favorable accuracy--speed trade-off. Under a shared budget, the set-based surrogate delivers higher joint success than the heuristic baselines and thus offers a stronger success--latency trade-off for real-time deployment. On the GPU benchmark, learned surrogates operate in the thousand-fold speedup regime over NSGA-II (Table~\ref{tab:runtime}), indicating that surrogate placement can meet real-time ITS decision loops under the stated assumptions. The success-rate curves and Pareto metrics further highlight that permutation-invariant modeling yields a more reliable coverage--SNR--count balance, while elongated or low-density scenes remain the dominant failure modes. These findings support a closed-loop deployment view in which placement is refreshed on triggers, projected to satisfy safety constraints, and backed by a fallback heuristic or short local search when confidence is insufficient.

Future work will incorporate temporal dynamics, UAV motion constraints, and energy budgets to move from static snapshots to closed-loop deployment. We also plan to strengthen the propagation model with interference and non-line-of-sight effects and validate the benchmark on additional datasets or field measurements.